\documentclass[a4paper,fleqn]{cas-sc}

\usepackage[numbers]{natbib}

\usepackage{tikz-dependency}
\usepackage{tikz}
\usepackage{tikz-qtree}
\usepackage{latexsym}
\usepackage{amssymb}
\usepackage{amsmath}
\usepackage{arydshln}

\def\tsc#1{\csdef{#1}{\textsc{\lowercase{#1}}\xspace}}
\tsc{WGM}
\tsc{QE}

\begin{document}
\let\WriteBookmarks\relax
\def\floatpagepagefraction{1}
\def\textpagefraction{.001}

\newcommand{\added}[1]{\textcolor{black}{#1}}
\newcommand{\new}[1]{\textcolor{black}{#1}}

\newcommand{\change}[1]{\textcolor{black}{#1}}
\newcommand{\spellchange}[1]{\textcolor{black}{#1}}

\shorttitle{Dependency Parsing with Bottom-up Hierarchical Pointer Networks}    

\shortauthors{D. Fern\'andez-Gonz\'alez, C. G\'omez-Rodr\'iguez.}  

\title [mode = title]{Dependency Parsing with Bottom-up Hierarchical Pointer Networks}  



%

\author[1]{Daniel Fern\'{a}ndez-Gonz\'{a}lez}[orcid=0000-0002-6733-2371]

\cormark[1]


\ead{d.fgonzalez@udc.es}

\ead[url]{https://danifg.github.io}

\credit{Conceptualization, methodology, software, validation, formal analysis, investigation, data curation, writing - original draft, writing - review \& editing, visualization}

\affiliation[1]{organization={Universidade da Coru\~{n}a, CITIC, FASTPARSE Lab, LyS Group, Depto. de Ciencias de la Computaci\'{o}n y Tecnolog\'{i}as de la Informaci\'{o}n},
            addressline={Campus de Elvi\~{n}a, s/n }, 
            city={A Coru\~{n}a},
            postcode={15071}, 
            country={Spain}}

\author[1]{Carlos G\'{o}mez-Rodr\'{i}guez}[orcid=0000-0003-0752-8812]


\ead{carlos.gomez@udc.es}

\ead[url]{http://www.grupolys.org/~cgomezr/}

\credit{Conceptualization, validation, formal analysis, writing - review \& editing, supervision, project administration, funding acquisition}


\cortext[1]{Corresponding author}



\begin{abstract}
Dependency parsing is a crucial step towards deep language understanding and, therefore, widely demanded by numerous Natural Language Processing applications. In particular,
left-to-right and top-down transition-based algorithms that rely on Pointer Networks are among the most accurate approaches for 
performing 
dependency parsing. Additionally, it has been observed for the top-down algorithm that Pointer Networks' sequential decoding can be improved by implementing a hierarchical variant, more adequate to model dependency structures. Considering all this, we develop a \spellchange{bottom-up oriented} Hierarchical Pointer Network 
for the left-to-right parser and propose two novel transition-based alternatives: an approach that parses a sentence in right-to-left order and a variant that \spellchange{does so} from the outside in.
We empirically test the proposed neural architecture with the different algorithms on a wide variety of languages, outperforming the original approach in 
practically
all of them and 
setting new state-of-the-art 
results on the English and Chinese Penn Treebanks for non-contextualized and BERT-based embeddings.
\end{abstract}



\maketitle

\section{Introduction}
\begin{figure}[h]
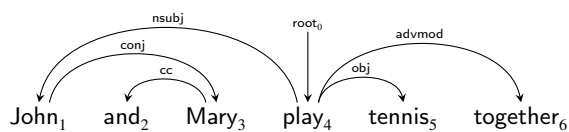

\centering
\begin{dependency}[theme = simple]
\begin{deptext}[column sep=1em]
John$_1$ \& and$_2$ \& Mary$_3$ \& play$_4$ \& tennis$_5$ \& together$_6$ \\
\end{deptext}
\deproot[edge unit distance=2ex]{4}{root$_0$}
\depedge{4}{1}{nsubj}
\depedge{3}{2}{cc}
\depedge{1}{3}{conj}
\depedge[edge unit distance=3ex]{4}{5}{obj}
\depedge[edge unit distance=3ex]{4}{6}{advmod}
\end{dependency}
\caption{Dependency tree for a given sentence.}
\label{fig:tree}
\end{figure}

\textit{Dependency parsing} consists in representing the grammatical structure of a given sentence by attaching each word to another (which will be considered its \textit{head} or \textit{parent}), to finally gather all these directed links into a dependency tree as the one included in Figure~\ref{fig:tree}. Additionally, these directed links or dependencies are enhanced with \textit{labels} that describe syntactic functions.

This syntactic information accurately provided by parsers \spellchange{in the form of} \textit{dependency trees} \spellchange{has proven} highly useful for a huge variety of Natural Language Processing (NLP) tasks.
In particular, dependency parsing has been recently used for machine translation 
\citep{ZHANG2021103427},
opinion mining \citep{SUN201710,zhang-etal-2020-syntax}, 
event extraction \citep{Nguyen2019}, question answering \citep{CAOPMID:31562071}, 
\change{sentiment analysis \citep{bai-etal-2021-syntax,barnes-etal-2021-structured,senticparser}}, coreference resolution \citep{SUKTHANKER2020139},
summarization \citep{balachandran-etal-2021-structsum} or semantic role labeling and named entity recognition \citep{sachan-etal-2021-syntax}, among others.

\textit{Pointer Networks} \citep{Vinyals15} have notably succeeded in implementing 
highly accurate versions of one of the most widely used dependency parsing paradigms: \textit{transition-based} dependency parsers.
In particular, two different algorithms have been proposed: a \textit{top-down} approach \citep{ma-etal-2018-stack} that, at each step and starting from the root node, connects each word to one of its children; and a \textit{left-to-right} variant \citep{fernandez-gonzalez-gomez-rodriguez-2019-left} 
that, starting from the left, attaches each word of the sentence to its parent. Apart from \spellchange{being twice as fast as} the former, the latter outperforms the top-down variant in terms of accuracy in practically all datasets tested so far.

As transition-based algorithms \citep{nivre-2003-efficient}, both perform the parsing process as a sequential decoding where, at each step or parsing state, all permissible actions (which do not violate the single-head or acyclicity constraints) are evaluated and the highest-scoring one is greedily applied, generating a new state. This sequential decoding differs from the approach followed by their main competitors: 
\textit{graph-based} algorithms \citep{mcdonald-etal-2005-online}. These previously score all possible arcs (or sets of arcs) and then, during decoding, search for the highest-scoring valid dependency tree. While recent graph-based models use a head-selection strategy \citep{Zhang17,DozatM17}
similar to that followed by the left-to-right transition-based parser \citep{fernandez-gonzalez-gomez-rodriguez-2019-left}, the training and decoding processes are completely different: at each decoding step, only permissible 
\spellchange{arc transitions}
are evaluated in the transition-based approach, and scores are computed taking into account a sequence of previous decisions encoded through the decoder. On the contrary, these simplified graph-based models independently score all possible parents for each word and then, during decoding, only the highest-scoring ones are kept regardless of the parents chosen for the other words. Then, at the end of the process, a maximum spanning tree algorithm is applied (if necessary)  to output a well-formed tree.\footnote{Additionally, while in transition-based parsers like \citep{ma-etal-2018-stack} and \citep{fernandez-gonzalez-gomez-rodriguez-2019-left}, the sequential decoding makes it possible to easily add high-order features and projectivity constraints without increasing runtime complexity, these enhancements in graph-based models lead to a performance penalty. 
} This difference can be also seen empirically, as the left-to-right transition-based parser achieves higher accuracies than the approach by \citet{DozatM17} in several widely-known benchmarks.

While this sequential decoding (typically implemented by a recurrent neural network) seems to be beneficial for 
dependency parsing, it might lead to accuracy losses due to error propagation: 
mistaken past decisions
will affect future actions, 
especially harming performance on long-range dependencies and attachments created in final steps. This limitation 
present in classic transition-based parsers also affects recent algorithms based on Pointer Networks, as decoder states located at the end of the sequence tend to forget relevant information from the past and are more prone to suffer from error propagation. 

In fact, \citet{liu-etal-2019-hierarchical} propose a \textit{hierarchical} decoding for the top-down algorithm \citep{ma-etal-2018-stack} by having access, at each step, to information about the focus word's parent and siblings created in the past, introducing not only knowledge about distant decoder states relevant for future decisions, but also an underlying tree structure to the decoding process (more appropriate for modelling dependency graphs).

In this paper, we initially develop a general Hierarchical Pointer Network architecture with a bottom-up structured decoding for the left-to-right transition-based algorithm
\citep{fernandez-gonzalez-gomez-rodriguez-2019-left}. This will allow the decoder to have access to information about partial structures created in the past and will help to make better future decisions. Alternatively to this algorithm, we design two novel \spellchange{bottom-up oriented} transition systems that can be easily implemented on the proposed neural model.

Finally, we empirically show that the presented architecture\footnote{Source code available at \url{https://github.com/danifg/BottomUp-Hierarchical-PtrNet}.}
with any transition-based algorithm 
provides improvements in accuracy on ten different languages from Universal Dependencies \citep{nivre-etal-2016-universal} and obtains state-of-the-art scores on the 
commonly-used
English and Chinese Penn Treebanks \citep{marcus93,Xue2005}.

The remainder of this article is organized as follows: Section~\ref{sec:preliminaries} introduces the baseline left-to-right parser by \citet{fernandez-gonzalez-gomez-rodriguez-2019-left} and briefly presents the top-down Hierarchical Pointer Network by \citet{liu-etal-2019-hierarchical}. In Section~\ref{sec:proposal}, we describe in detail the proposed bottom-up Hierarchical Pointer Network and how it is adapted to the left-to-right algorithm. Section~\ref{sec:newalgos} presents new transition systems and how they are implemented on the novel neural architecture. In Section~\ref{sec:experiments}, we extensively evaluate the proposed neural model with each parsing strategy on numerous datasets, as well as include a thorough analysis of their performance. Lastly, Section~\ref{sec:conclusion} contains a final discussion.

\section{Preliminaries}
\label{sec:preliminaries}
\subsection{Left-to-Right Transition System}
\begin{figure}[h]
\centering
\includegraphics[width=0.49\textwidth]{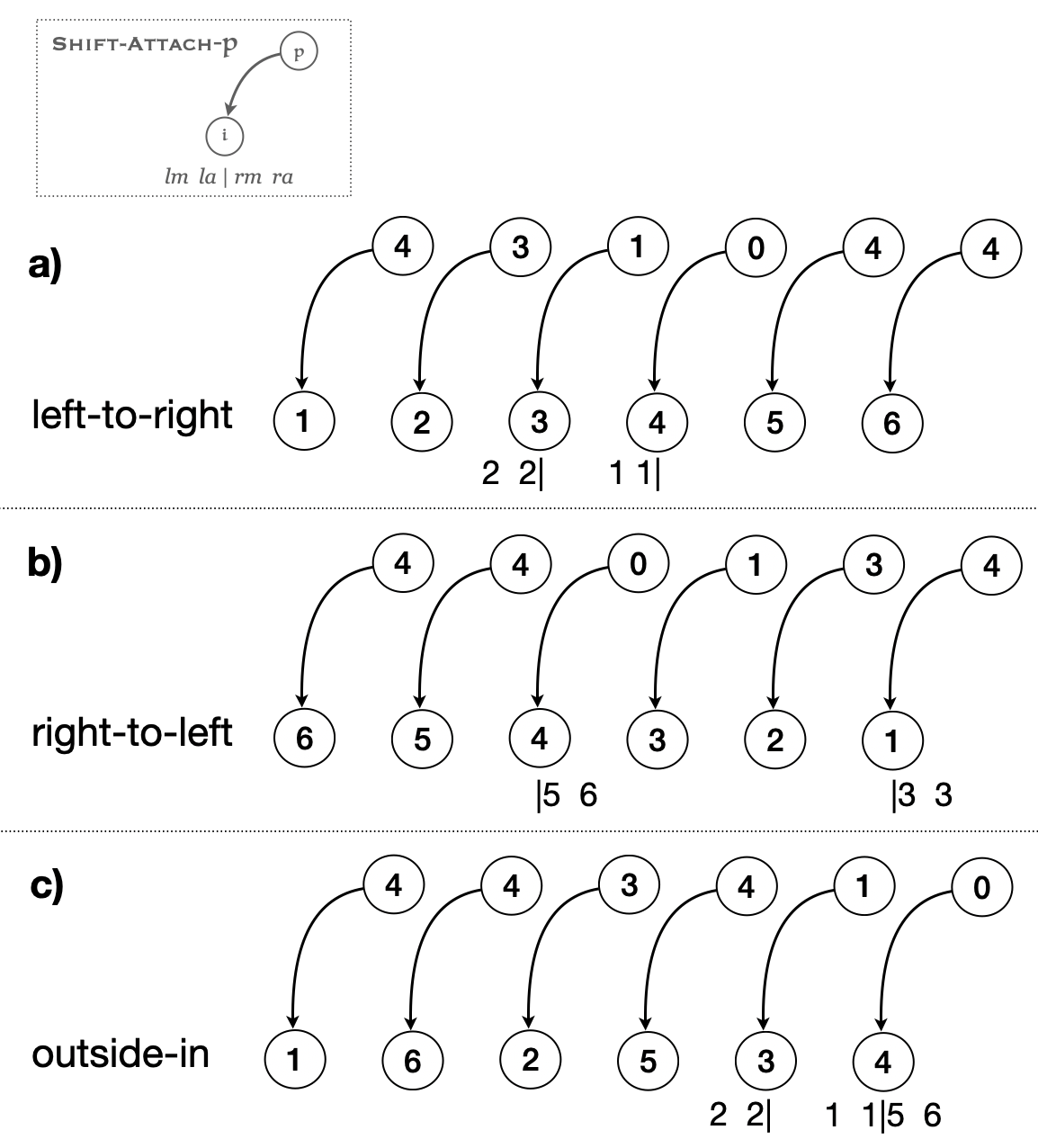}
\caption{Attachment order and leftmost (\texttt{lm}), rightmost (\texttt{rm}), last-attached left (\texttt{la}) and right (\texttt{ra}) dependents that are available at each decoding step when parsing the sentence in Figure~\ref{fig:tree} with the three available transition-based algorithms. Note that words $w_i$ are represented by the positional index $i$.}
\label{fig:algos}
\end{figure}

\added{\citet{fernandez-gonzalez-gomez-rodriguez-2019-left} propose an efficient left-to-right transition system that is defined by a \textit{focus word pointer} $i$, which is used to point to the word currently being processed $w_i$, and a single \textsc{Shift-Attach}-$p$ transition, which assigns a parent word $w_p$ (in position $p$) to $w_i$ (producing the dependency arc $w_p \rightarrow w_i$) and then \spellchange{moves} the pointer \spellchange{making it point to the} next token $w_{i+1}$. Starting at the beginning of a sentence (of length $n$), this algorithm 
sequentially 
attaches each token $w_i$ to its parent in just $n$ steps.}\footnote{The top-down transition system by \citet{ma-etal-2018-stack} needs 2$n$-1 steps to parse a sentence of length $n$.} In Figure~\ref{fig:algos}(a), we depict the sequential prediction of  \textsc{Shift-Attach}-$p$ transitions that produces all arcs in the dependency tree in Figure~\ref{fig:tree} following a left-to-right transition system.

In order to incrementally build a well-formed dependency tree during decoding, only \textsc{Shift-Attach}-$p$ transitions that do not generate cycles in the already-built dependency graph are allowed. Additionally, the left-to-right transition system can efficiently produce unrestricted dependency graphs, including non-projective structures.\footnote{\added{Non-projective dependency graphs are harder to produce since they are able to represent more complex syntactic phenomena by allowing crossing arcs.}} However, the projectivity constraint can \spellchange{optionally} be enforced \spellchange{by} discarding transitions that produce crossing dependencies\change{, which can be useful} \spellchange{in treebanks} with a negligible presence of non-projectivity \citep{multipointer}.

\subsection{Pointer Networks for Left-to-Right Dependency Parsing}
\label{sec:l2r}
The left-to-right algorithm can be easily implemented by a Pointer Network \citep{Vinyals15} to \added{undertake} non-projective dependency parsing. \change{Pointer Networks are one among the various recent neural models that seek to improve performance on various Artificial Intelligence tasks \citep{10.3389/fnins.2021.601109,10.3389/fnins.2022.850945}.} \added{These sequence-to-sequence neural networks are trained to output a sequence of positions (discrete numbers) from an input sequence of items.} In our case, the output is the sequence $\mathbf{p} = p_1, p_2, \dots, p_n$
of 
values 
necessary to parameterize the \textsc{Shift-Attach}-$p$ transition and build a dependency graph for the input sentence $\mathbf{w} = w_1, \dots ,w_n$. These positions (or values of $p$ in the left-to-right transition system) are selected by \added{an attention mechanism} \citep{Bahdanau2014} over the input sentence. This avoids having to keep a \added{fixed output dictionary size}, which will vary for each input sentence with respect to its length. 

More specifically, the encoder-decoder architecture of the original left-to-right parser 
\citep{fernandez-gonzalez-gomez-rodriguez-2019-left} is designed as follows:
\paragraph{Encoder} Given an input sentence $\mathbf{w} = w_1, \dots ,w_n$, each word $w_i$ is initially represented as $x_i$: i.e., a concatenation of character-level representations ($\mathbf{e}^c_i$) (extracted by max-pooling-based Convolutional Neural Networks \citep{Ma2016}), POS tag embeddings ($\mathbf{e}^p_i$) and word embeddings ($\mathbf{e}^w_i$): 
$$\mathbf{x}_i = \mathbf{e}^c_i \oplus \mathbf{e}^w_i \oplus \mathbf{e}^p_i$$
\noindent Then, each vector representation $\mathbf{x}_i$ is \added{passed through}
a BiLSTM \change{(\textit{Bidirectional Long Short-Term Memory})}  to generate the encoder hidden state $\mathbf{h}_i$.
$$ \mathbf{h}_i = \mathbf{h}_{li}\oplus\mathbf{h}_{ri} = \mathbf{BiLSTM}(\mathbf{x}_i)$$

\noindent \added{where $\mathbf{h}_{li}$ and $\mathbf{h}_{ri}$ are the outputs of a leftward and a rightward uni-directional LSTM \change{(\textit{Long Short-Term Memory})}, respectively. This results in the sequence $\mathbf{h}_0 \dots \mathbf{h}_n$ that encodes the input sentence (\spellchange{where $\mathbf{h}_0$ is} a special  representation for the artificial ROOT node).}

We also report accuracies with our encoder enhanced with 
contextualized word embeddings extracted from the widely-used language model BERT \citep{devlin-etal-2019-bert}. In those cases, BERT-based word embeddings are directly concatenated to the 
basic word representation $\mathbf{x}_i$ before passing it through the BiLSTM:
$$\mathbf{x}'_i = \mathbf{x}_i \oplus \mathbf{e}^{BERT}_i$$
$$\mathbf{h}_i = \mathbf{BiLSTM}(\mathbf{x}'_i)$$

\paragraph{Decoder} The sequential decoding that models the transition-based behaviour is implemented by a unidirectional LSTM, \added{which is in charge of generating a new \textit{decoder hidden state} $\mathbf{s}_t$ at each time step $t$. As input it 
receives the encoder hidden state $\mathbf{h}_i$ (representation of the current focus word $w_i$) and, as a standard 
recurrent neural network, it is also conditioned by the previous decoder state $\mathbf{s}_{t-1}$:}
$$\mathbf{s}_t = \mathbf{LSTM}(\mathbf{h}_i) = f(\mathbf{s}_{t-1},\mathbf{h}_i)$$
\noindent In the original work, authors added to $\mathbf{h}_i$  the previous and next encoder hidden states as extra features; however, in a followup paper, they removed those additions since the bidirectional LSTM is already encoding such information and, therefore, they do not lead to significant improvements \citep{multipointer}. 

The resulting decoder state $\mathbf{s}_t$ (which represents the current focus word plus the past decisions made so far) is used for computing scores of all possible words $w_j$ from the input sentence (encoded as vectors $\mathbf{h}_j$ with $j \in [0,n]$ and $j \neq i$) as parent of $w_i$. These scores are obtained by a biaffine scoring function \citep{DozatM17} to finally compute the attention distribution in \spellchange{a} vector $\mathbf{a}_t$:
$$\mathbf{v}_{tj} = \mathbf{score}(\mathbf{s}_t, \mathbf{h}_j)= f_1(\mathbf{s}_t)^T \mathbf{W} f_2(\mathbf{h}_j)
+\mathbf{U}^Tf_1(\mathbf{s}_t) + \mathbf{V}^Tf_2(\mathbf{h}_j) + \mathbf{b};$$
\noindent $$\mathbf{a}_t = \mathbf{softmax}(\mathbf{v}_t)$$
where $\mathbf{W}$, $\mathbf{U}$ and $\mathbf{V}$ are the weights and, \added{as observed by \citet{DozatM17}, $f_1(\cdot)$ and $f_2(\cdot)$ are 
multilayer perceptrons (MLP) for reducing dimensionality and avoiding model overfitting.} 

Then, \spellchange{the} attention vector $\mathbf{a}_t$ is employed for implementing a pointer over the input sentence, selecting, at each step $t$, the highest-scoring position (where the parent word $w_p$ is located)  and, therefore, providing a value $p$ necessary \spellchange{to apply} a \textsc{Shift-Attach}-$p$ transition and \spellchange{build} the arc $w_p \rightarrow w_i$. Finally, as proposed by \citep{ma-etal-2018-stack}, \added{a multi-class classifier (based on the approach developed by \citet{DozatM17})} is separately trained to predict the arc label of each dependency created by the pointer.

\added{Since the decoder performs $n$ attachments to process an input sentence with length $n$ \spellchange{and, additionally, attention is computed at each step} over the whole sentence, the time complexity of the parsing process is $O(n^2)$.}


\subsection{Top-down Hierarchical Pointer Networks}
\citet{liu-etal-2019-hierarchical} introduced Hierarchical Pointer Networks for the top-down transition-based algorithm developed by \citet{ma-etal-2018-stack}. The top-down transition system consists of two actions and a stack: one transition for connecting the word on top of the stack to one of its children and push it into the stack, and another action for popping the current focus word on top. 
\citet{liu-etal-2019-hierarchical} design a structured decoding for the original Pointer Network, where not only the immediately-previous decoder state $\mathbf{s}_{t-1}$ of the LSTM is considered for choosing the next action to be applied on the current word on top,
but also decoder hidden states generated when the current focus word's parent 
and last sibling 
were assigned following the top-down transition system. More graphically, in Figure~\ref{fig:tree}, when, for instance, the current focus word to be processed is \textit{John}, decoder hidden states of its parent \textit{play} and last-assigned sibling \textit{together}\footnote{Assuming that in the top-down transition system, right children are attached first in an inside-out order.} are 
\spellchange{taken into consideration when choosing}
\textit{John}'s children. With this strategy, \spellchange{the authors} manage to introduce an explicit structural inductive bias into the original linear decoder, allowing a more adequate modelling for the generation of dependency graphs. \citet{liu-etal-2019-hierarchical} also empirically show that their top-down Hierarchical Pointer Networks are beneficial \spellchange{for the accuracy of} arcs created in final steps, which is especially crucial on long sentences.

\section{Bottom-up Hierarchical Pointer Networks}
\label{sec:proposal}
Based on the research work by \citet{liu-etal-2019-hierarchical}, we propose a Hierarchical Pointer Network for the left-to-right transition-based algorithm 
\citep{fernandez-gonzalez-gomez-rodriguez-2019-left} and other related parsing strategies where dependents of the focus word may already have been attached (as we will design more such algorithms later). In particular, instead of parent or sibling information, we design a bottom-up structured tree decoding by considering decoder hidden states of the focus word's relevant dependents. More concretely, to alleviate the gradual loss of relevant information \spellchange{during sequential} decoding, we propose to consider decoder hidden states of the leftmost and rightmost dependents already attached to the current focus word, as well as the most recently attached left and right dependents. Note that not all this information is available in all strategies, e.g., the left-to-right algorithm has no access to right dependencies of the focus word.

All this will not only provide valuable knowledge about words processed in the past (especially those attached by long-range dependencies), but will also keep an underlying tree structure that will 
help the decoder to make good decisions throughout the parsing process. For instance, when the sentence from Figure~\ref{fig:tree} is parsed following a purely bottom-up strategy\footnote{Please note that the left-to-right transition system is not purely bottom-up since, when a word is attached to its parent, \spellchange{not all of its children have necessarily been assigned yet}.} and \spellchange{the} current focus word \textit{play} must be attached to its parent, it would be helpful to have access to information about its leftmost and rightmost dependents (\textit{John} and \textit{together}, respectively), as well as the \spellchange{most} \spellchange{recently attached} right dependent (\textit{tennis}). 

\subsection{Structured Tree Decoder}
To implement this novel bottom-up Hierarchical Pointer Network variant, we keep the same encoder as the original approach  \citep{fernandez-gonzalez-gomez-rodriguez-2019-left}, but use a bottom-up hierarchical decoder instead. 

More in detail, at each step $t$ with current focus word $w_i$, the generated decoder hidden state  $\mathbf{s}_t$ is directly conditioned, apart from by the previous decoder state $\mathbf{s}_{t-1}$ and the encoder hidden state $\mathbf{h}_i$, by the decoder states of the already-attached leftmost ($\mathbf{s}_{lm(t)}$) and rightmost ($\mathbf{s}_{rm(t)}$) dependents of $w_i$, as well as the decoder states of the left ($\mathbf{s}_{la(t)}$) and right ($\mathbf{s}_{ra(t)}$) last-attached dependents:\footnote{Note that $\mathbf{s}_{lm(t)}$ and $\mathbf{s}_{rm(t)}$ might have the same values as $\mathbf{s}_{la(t)}$ and $\mathbf{s}_{ra(t)}$, respectively, when just one left or right dependent were assigned for the current focus word.} 
$$\mathbf{s}_t = f(\mathbf{s}_{lm(t)},\mathbf{s}_{rm(t)},\mathbf{s}_{la(t)},\mathbf{s}_{ra(t)},\mathbf{s}_{t-1},\mathbf{h}_i)$$
\noindent where $f(\cdot)$ is a \textit{fusion function} that combines all components into a single decoder state.

Following \citet{liu-etal-2019-hierarchical}, we do not directly feed these six components to the decoder, but utilize a \textit{gating mechanism} to allow our model to adequately extract the most useful information at each decoding step. In particular, we implement two different gating functions:
\begin{equation}
\tag{\textsc{Gate1}}
\centering
\mathbf{g}_t = \mathbf{sigmoid}(\mathbf{W}_{gp}\mathbf{s}_{t-1}
+ \mathbf{W}_{glm}\mathbf{s}_{lm(t)}
+ \mathbf{W}_{grm}\mathbf{s}_{rm(t)}
+ \mathbf{W}_{gla}\mathbf{s}_{la(t)}
+ \mathbf{W}_{gra}\mathbf{s}_{ra(t)} + \mathbf{b}_g)
\end{equation}
\begin{equation}
\tag{\textsc{Gate2}}
\mathbf{g}_t = \mathbf{sigmoid}(\mathbf{W}_{glm}(\mathbf{s}_{t-1} \odot \mathbf{s}_{lm(t)})
+ \mathbf{W}_{grm}(\mathbf{s}_{t-1} \odot \mathbf{s}_{rm(t)})
+ \mathbf{W}_{gla}(\mathbf{s}_{t-1} \odot \mathbf{s}_{la(t)})
+ \mathbf{W}_{gra}(\mathbf{s}_{t-1} \odot \mathbf{s}_{ra(t)}) + \mathbf{b}_g)
\end{equation}
\noindent where $\mathbf{W}_{gp}$, $\mathbf{W}_{glm}$, $\mathbf{W}_{grm}$, $\mathbf{W}_{gla}$, $\mathbf{W}_{gra}$ and $\mathbf{b}_g$ are gating weights. While all decoder states are equally weighted in \textsc{Gate1}, the element-wise product used in \textsc{Gate2} has the effect of similarity comparison among the previous decoder state $\mathbf{s}_{t-1}$ and the other components.

These gating functions are then used to define the fusion function $f$ as follows:
$$\mathbf{h}_t' = tanh(\mathbf{W}_{p}\mathbf{s}_{t-1} + \mathbf{W}_{lm}\mathbf{s}_{lm(t)}
+ \mathbf{W}_{rm}\mathbf{s}_{rm(t)}
+ \mathbf{W}_{la}\mathbf{s}_{la(t)}
+ \mathbf{W}_{ra}\mathbf{s}_{ra(t)});$$
$$\mathbf{h}_t'' = \mathbf{g}_t \odot \mathbf{h}_t';$$
$$\mathbf{s}_t = \mathbf{LSTM}(\mathbf{h}_t'',\mathbf{h}_i)$$
\noindent where $\mathbf{W}_{lm}$, $\mathbf{W}_{rm}$, $\mathbf{W}_{la}$ and $\mathbf{W}_{ra}$ are the weights for combining the five different decoder states into one intermediate hidden state $\mathbf{h}_t'$. After that, the gating mechanism $\mathbf{g}_t$ is applied to control the information flow and generate the hidden state $\mathbf{h}_t''$, which will be fed into the decoder together with the encoder hidden state $\mathbf{h}_i$. 

As shown in Section~\ref{sec:l2r}, the resulting decoder state $\mathbf{s}_t$ is then used for computing the attention vector $\mathbf{a}_t$
that will work as a pointer over the input sentence.

\subsection{Model Specifics for the Left-to-Right Transition System}
While the left-to-right transition system can be directly implemented on the described Hierarchical Pointer Network, it cannot use its full potential. This transition-based approach does not follow a fully bottom-up strategy during the left-to-right decoding\footnote{Left dependents are added bottom–up and right dependents top–down.} and, therefore, right dependents are not available when they are needed. In the example, when the word \textit{play} is under processing, only the leftmost dependent \textit{John} is available since no right dependents \spellchange{have been} created yet and, when the words \textit{tennis} and \textit{together} \spellchange{are finally} attached, they are no longer needed since the word \textit{play} will have already been processed. This led us to simplify the definition of the gating-based fusion function $f$ for the left-to-right transition system by removing decoder states of right dependents:
\begin{equation}
\tag{l-adapted}
 \mathbf{s}_t = f(\mathbf{s}_{lm(t)},\mathbf{s}_{la(t)},\mathbf{s}_{t-1},\mathbf{h}_i)   
\end{equation}

\noindent Since, in some languages, it might be the case that words have either only one left dependent ($\mathbf{s}_{lm(t)}$ and $\mathbf{s}_{la(t)}$ being the same) or the left last-attached dependent was created in the previous time step ($\mathbf{s}_{t-1}$ and $\mathbf{s}_{la(t)}$ being the same), we also experiment with a simpler variant of $f$ that just considers decoder states $\mathbf{s}_{t-1}$ and $\mathbf{s}_{lm(t)}$:
\begin{equation}
\tag{l-simple}
 \mathbf{s}_t = f(\mathbf{s}_{lm(t)},\mathbf{s}_{t-1},\mathbf{h}_i)   
\end{equation}

Finally, it is worth mentioning that the hierarchical decoder does not penalize the quadratic runtime complexity of the left-to-right parser and, as the original work \citep{fernandez-gonzalez-gomez-rodriguez-2019-left}, it is trained by minimizing the total log loss (cross entropy) for choosing the correct sequence of \textsc{Shift-Attach}-$p$ transitions to build a gold dependency tree for the input sentence $\mathbf{w}$ (i.e. predicting the correct sequence of indices $p$, with each decision ($p_t$) being conditioned by previous ones ($p_{<t}$)):
$$\mathcal{L}(\theta) = - \sum_{t=1}^{T} log P_\theta (p_t | p_{<t},\mathbf{w})$$

\noindent  By optimizing the sum of their objectives, we simultaneously train the pointer and the labeler.

\begin{figure*}
\centering
\includegraphics[width=0.8\textwidth]{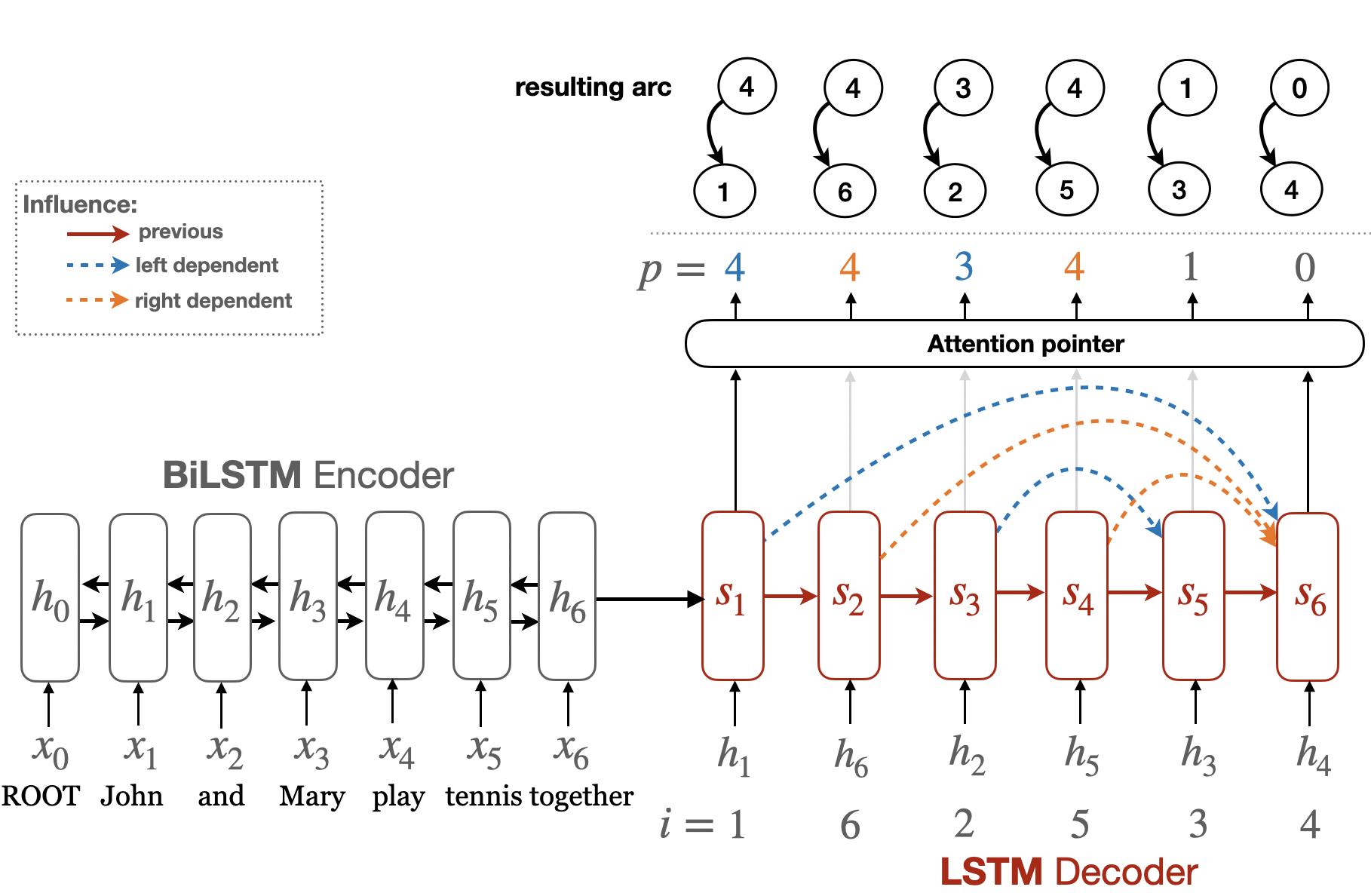}
\caption{Neural architecture for the Hierarchical Pointer Network and decoding steps necessary to produce \spellchange{the} dependency tree in Figure~\ref{fig:tree} with the outside-in transition system. Note that some dependent influences, such as the one generated by arc $1 \rightarrow 3$ in decoder state $\mathbf{s_5}$, will not be used as the word in position $1$ was already processed.}
\label{fig:network}
\end{figure*}

\section{Alternative Transition-based Algorithms}
\label{sec:newalgos}
Apart from the existing left-to-right algorithm, other transition systems can be implemented on the proposed bottom-up Hierarchical Pointer Network. In particular, we develop two alternative approaches to sequentially parse a sentence by attaching each word to its parent.

\paragraph{Right-to-Left}
 While a left-to-right hierarchical decoding might be more adequate for modelling 
 left-branching 
 languages where long-range dependencies tend to be leftward arcs (such as Turkish and Korean), for right-branching 
 languages with a high predominance of long rightward arcs (such as Arabic and Hebrew),
 a right-to-left transition system could be a perfect fit.\footnote{Please note that, following common usage in the parsing literature, naming conventions in this paper use the terms ``left'' and ``right'' to refer to temporal order (under the convention that a word is to the ``left'' of another if it comes first), not to spatial order when the words are written in a given script. This means that in languages with a right-to-left writing system, such as Arabic and Hebrew, the original left-to-right parser will begin reading each line of text from the right (i.e., the beginning), and the right-to-left algorithm will begin from the left (i.e., the end). Similarly, the graphical representation of what we call a ``leftward arc'' would be an arrow pointing to the right in those languages.}
This parses a sentence starting from the last word $w_n$ and it provides a \textsc{Shift-Attach}-$p$ transition that, at each step, assigns a parent to the current focus word $w_i$ and moves $i$ to point to word $w_{i-1}$. In Figure~\ref{fig:algos}(b), we describe the order in which arcs are created with the right-to-left transition system to generate the dependency tree in Figure~\ref{fig:tree} and the available dependents at each decoding step. Symmetrically to the left-to-right algorithm, we adjust the fusion function $f$ to process only decoder states of right dependents:
\begin{equation}
\tag{r-adapted}
\mathbf{s}_t = f(\mathbf{s}_{rm(t)},\mathbf{s}_{ra(t)},\mathbf{s}_{t-1},\mathbf{h}_i)
\end{equation}
\noindent Additionally, following the same reasoning as for the left-to-right algorithm, we also implement a simplified alternative that removes $\mathbf{s}_{ra(t)}$ from the equation:
\begin{equation}
\tag{r-simple}
\mathbf{s}_t = f(\mathbf{s}_{rm(t)},\mathbf{s}_{t-1},\mathbf{h}_i)
\end{equation}

\paragraph{Outside-in}
In order to fully use the proposed neural architecture and, therefore, have access to left and right dependents during the decoding process, we design an 
\textit{outside-in} transition system: it parses a sentence starting from the leftmost word ($w_1$) and continuing with the rightmost word ($w_n$) towards the middle of the sentence, alternating words from the left and the right. To that end, a canonical order $w_1, w_n, w_2, w_{n-1}, \dots , w_{\lfloor \frac{n+1}{2} \rfloor}$ is determined at the \spellchange{beginning. At} each decoding step the \textsc{Shift-Attach}-$p$ transition, apart from attaching $w_i$ to its parent, moves $i$ according to that order.  In Figure~\ref{fig:algos}(c), we can see an example of how this outside-in algorithm works and how left and right dependents are available during the decoding process. Since this strategy allows us to use decoder states of left and right dependents, the fusion function $f$ can be fully used. For the same reasons as stated for the other transition systems, we also propose a simplification that discards last-attached dependents as follows:
\begin{equation}
\tag{simple}
\mathbf{s}_t = f(\mathbf{s}_{lm(t)},\mathbf{s}_{rm(t)},\mathbf{s}_{t-1},\mathbf{h}_i)
\end{equation}
\noindent Note that both alternative transition systems are guaranteed to parse a sentence of length $n$ in just $n$ steps, keeping the same runtime complexity as the left-to-right algorithm, and are likewise trained, except for the order in which transitions (values of $p$) are predicted.

Figure~\ref{fig:network} depicts the proposed Hierarchical Pointer Network 
and presents the decoding procedure for producing the dependency tree described in Figure~\ref{fig:tree} with the 
outside-in transition system. In this sketch, it can be graphically seen how decoder hidden states of dependents influence the decoding process.

Finally, it is worth mentioning that
it is not possible to design a purely bottom-up transition system on the Pointer Network framework using $n$ transitions (one per word)
as, regardless of the order in which words are considered, it is not possible to guarantee that dependents are always processed before heads (e.g., the first node considered will not necessarily be a leaf).

\section{Experiments}
\label{sec:experiments}
\subsection{Data}
We conduct experiments on a wide variety of languages from Universal Dependencies v2.6 \citep{nivre-etal-2016-universal}. Following \citep{kulmizev-etal-2019-deep}, we choose ten treebanks 
from different language families, with different morphological complexity and with different predominances of long-range dependencies. These are detailed in Table~\ref{tab:treebanks}. 

\begin{table}[h]
\begin{small}
\begin{center}
\centering
\begin{tabular}{@{\hskip 0pt}llccccl@{\hskip 0pt}}
\toprule
\scriptsize{\textbf{Language}} & \scriptsize{\textbf{Treebank}} & \scriptsize{\textbf{Family}} & \scriptsize{\textbf{Order}} & \scriptsize{\textbf{Size}} & \scriptsize{\textbf{\%long}} & \scriptsize{\textbf{\%left}} \\
\midrule
Arabic  & PADT  &  AA & VSO  & 6.1k & 16.62 & 10.30 \\
Basque  & BDT  &  LI & SOV  & 5.4k & 
20.17 & 	44.46 
 \\
Chinese  & GSD  &  ST & SVO  & 4.0k & 
25.31 &	59.44
 \\
English  & EWT  &  IE & SVO  & 12.5k & 
16.43 &	25.49
 \\
Finnish  & TDT  &  UR & SVO  & 12.2k & 
17.45 &	26.83
 \\
Hebrew  & HTB  &  AA & SVO  & 5.2k & 
17.82 & 	20.60
 \\
Italian  & ISDT  &  IE & SVO  & 13.1k & 
16.20 & 	22.69
 \\
Korean  & GSD  &  KO & SOV  & 4.4k & 
14.76 &	78.40
 \\
Swedish  & Talbanken  &  IE & SVO  & 4.3k & 
19.22 &	29.51
 \\
Turkish  & IMST  &  TU & SOV  & 3.7k & 
14.84 & 	70.69
\\
\bottomrule
\end{tabular}
\centering
\setlength{\abovecaptionskip}{4pt}
\caption{\spellchange{Details of the treebanks} used in our experiments. Family = Afro-Asiatic (AA), Indo-European (IE), Koreanic (KO), Language isolate (LI), Sino-Tibetan (ST), Turkic (TU) or Uralic (UR). Order = dominant word order according to WALS \citep{haspelmath2005world}. Size = number of training sentences. \%long = percentage of long arcs (length $>$ 4) in the dev split. \%left = percentage of long arcs that are leftward. }
\label{tab:treebanks}
\end{center}
\end{small}
\end{table}

Additionally, we evaluate the proposed neural architecture on two widely-known benchmarks: 
the English Penn Treebank (PTB) \citep{marcus93} version with Stanford Dependencies \citep{de-marneffe-manning-2008-stanford}
and, following \citep{multipointer}, without any PoS tags;\footnote{It has been shown that using predicted PoS tags does not lead to  accuracy gains in parsers built on Pointer Networks \citep{ma-etal-2018-stack}.} 
and the dependency conversion \citep{zhang-clark-2008-tale} of the Chinese Penn Treebank 5.1 (CTB) \citep{Xue2005} with gold POS tags.

\added{Since random initializations are performed, we report the average Labelled and Unlabelled Attachment Scores (LAS and UAS) over 3 repetitions for each experiment; and, during evaluation, we just exclude punctuation on PTB and CTB following standard practice.}

\subsection{Settings}
We use the code of the original parser by
\citet{fernandez-gonzalez-gomez-rodriguez-2019-left}
(with the modifications proposed for the single-task parser in \citep{multipointer}) and implement the bottom-up Hierarchical Pointer Network plus the two novel transition systems, allowing a homogeneous comparison against the baseline. Similarly to \citep{liu-etal-2019-hierarchical}, we experiment with the two available gate mechanisms in combination with the proposed transition systems and fusion function implementations. \change{All models were executed on an Intel(R) Core(TM) i7-8700K CPU @ 3.70GHz with two 12 GB GeForce GTX 1080 Ti GPUs.}

\added{While character and PoS tag embeddings are randomly initialized, we use pre-trained embeddings for initializing word vectors: structured-skipgram embeddings \citep{Ling2015} for English and Chinese, and Polyglot embeddings \citep{al-rfou-etal-2013-polyglot} for other languages. All embeddings are fine-tuned during training. }

For PTB and CTB, we additionally concatenate 
contextualized word embeddings obtained from 
the pre-trained language model 
BERT \citep{devlin-etal-2019-bert}. We follow the greener and less resource-consuming 
approach undertaken by \citep{multipointer} and directly feed 
BERT fixed
weights 
as described in Section~\ref{sec:l2r},
without any fine-tuning to our specific task. More specifically, we extract a combination of weights from \spellchange{layers 17-20} of BERT$_\textsc{LARGE}$ for PTB and weights from the 
\spellchange{11th} layer of BERT$_\textsc{BASE}$ for CTB, averaging BERT-based embeddings when a word is tokenized into more than one subword.

\added{Finally, we use beam size 10 for PTB and CTB, and 1 for \change{Universal Dependencies treebanks}; we employ the Adam optimizer \citep{Adam} for parameter optimization and further details about hyper-parameter selection are reported
in Table~\ref{tab:hyper}. }

\begin{table}[h]
\begin{footnotesize}
\centering
\begin{tabular}{@{\hskip 0pt}lc@{\hskip 0pt}}
\toprule
\textbf{Adam optimizer} &\\
\midrule
Batch size & 32 \\
$\beta_1$ & 0.9 \\
$\beta_2$ & 0.9 \\
Initial learning rate & 0.001 \\
Gradient clipping & 5.0 \\
Decay rate & 0.75 \\
\midrule
\textbf{Architecture} & \\
\midrule
BiLSTM size & 512 \\
BiLSTM number of layers & 3 \\
LSTM size & 512 \\
LSTM number of layers & 1 \\ 
LSTM layers dropout & 0.33 \\
CNN window size & 3 \\
CNN number of filters & 50 \\
MLP activation function & ELU \\
MLP number of layers & 1 \\
Arc MLP size & 512 \\ 
Label MLP size & 128 \\
UNK replacement probability & 0.5 \\
Character embedding dimension & 100\\
POS tag embedding dimension & 100\\
Word embedding dimension & 100\\
Embeddings dropout & 0.33 \\
English BERT embedding dimension & 1024\\
Chinese BERT embedding dimension & 768\\
\bottomrule
\end{tabular}
\setlength{\abovecaptionskip}{4pt}
\caption{Hyper-parameter selection.}
\label{tab:hyper}
\end{footnotesize}
\end{table}

\begin{table*}[tbp]
\begin{small}
\begin{center}
\centering
\begin{tabular}{@{\hskip 1pt}l@{\hskip 4pt}c@{\hskip 4pt}c@{\hskip 4pt}|cccccccccc|c@{\hskip 1pt}}
\toprule
\textbf{tran.} & \textbf{$f$} & \textbf{gate} & \textbf{ar} & \textbf{en} & \textbf{eu} & \textbf{fi} & \textbf{he} & \textbf{it} & \textbf{ko} & \textbf{sv} & \textbf{tr} & \textbf{zh} & \textbf{Avg.}\\
\hline
\multicolumn{14}{c}{\textbf{dev}} \\
\hline
\textbf{L2R} & - & - & 83.82 & 90.29 & 83.80 & 88.46 & 89.11 & 92.52 & 83.29 & 86.78 & 64.75 & 82.54 & 84.54 \\
\hline
\textbf{L2R} & l-adapted & \textsc{Gate1} & 84.15 & 90.51 & 84.67 & \textbf{88.97} & 89.27 & \textbf{92.63} & 83.81 & 87.41 & 65.35 & 83.62 & 85.04 \\ 
 & l-adapted & \textsc{Gate2} & 84.05 & 90.44 & 84.53 & 88.95 & 89.22 & 92.60 & 83.79 & 87.32 & 65.24  & 83.48 & 84.96 \\
 & l-simple & \textsc{Gate1} & 84.06 &  90.49 & 84.84 & 88.90  & 89.30 & 92.59 & 83.84 & \textbf{87.44} & \textbf{65.40} & 83.65 & \textbf{85.05} \\
 & l-simple & \textsc{Gate2} & 84.05 & 90.29  & 84.77 & 88.89 & 89.20 & 92.62 & \textbf{83.86} & 87.43 & 65.12 & 83.57 & 84.98 \\
\hline
\textbf{R2L} & r-adapted & \textsc{Gate1} & 84.21 & 90.41 & 84.83 & 88.95 & 89.21 & 92.47 & 83.66 & 87.28  & 65.25 & 83.50 & 84.98 \\
 & r-adapted & \textsc{Gate2} & 84.20 & 90.26 & \textbf{84.93} & 88.81 & 89.21 & 92.52 & 83.68 & 87.14 & 65.09 & 83.53 & 84.94 \\
 & r-simple & \textsc{Gate1} & \textbf{84.28} & \textbf{90.52} & 84.75 & 88.86 & 89.28 & 92.56 & 83.71 & 87.41 & 65.33 & \textbf{83.69} & \textbf{85.04} \\
 & r-simple & \textsc{Gate2} & 84.26 & 90.46 & 84.88 & 88.85 & 89.20 & 92.40 & 83.69 & 87.26 & 65.36 & 83.52 & 84.99 \\
 \hline
\textbf{O-I} & full & \textsc{Gate1} & 84.07 & 90.43 & 84.59 & 88.74 & 89.28 & 92.51 & 83.46  & 86.91 & 65.14 & 83.54 & 84.87 \\
 & full & \textsc{Gate2} & 84.00 & 90.33 & 84.42 & 88.80 & 89.09 & 92.38 & 83.45 & 86.96 & 65.07  & 83.47 & 84.80 \\
 & simple & \textsc{Gate1} & 84.13 & 90.48 & 84.54 & 88.77 & \textbf{89.36} & 92.55 & 83.56 & 87.06 & 65.12 & 83.49 & \textbf{84.91} \\
 & simple & \textsc{Gate2} & 84.04 & 90.35 & 84.53 & 88.77 & 89.10 & 92.33 & 83.53 & 87.07 & 65.14 & 83.28 & 84.81 \\
\hline
\multicolumn{14}{c}{\textbf{test}} \\
\hline
\textbf{L2R} & - & - & 84.20 & 89.32 & 84.53 & 88.45 & 87.75 & 92.46 & 86.01 & 88.92 & 66.12 & 83.68 & 85.14 \\
\textbf{L2R} & l-simple & \textsc{Gate1} & 84.45 & \textbf{89.43} & 85.24 & \textbf{89.00} & 88.23 & 92.50 & \textbf{86.21} & 89.16 & 66.63 & \textbf{84.99} & 85.58 \\
\textbf{R2L}  & r-simple & \textsc{Gate1} & \textbf{84.51} & \textbf{89.43} & \textbf{85.46} & 88.77 & 88.26 & \textbf{92.62} & 86.16 & \textbf{89.17} & \textbf{67.05} & 84.75 & \textbf{85.62} \\
\textbf{O-I}  & simple & \textsc{Gate1} & 84.31 & 89.33 & 85.33 & 88.88 & \textbf{88.27} & 92.54 & 86.14 & 88.99 & 66.60 & 84.43 & 85.48 \\
\hline
\end{tabular}
\centering
\setlength{\abovecaptionskip}{4pt}
\caption{LAS comparison of the 
original left-to-right parser with sequential decoding against the \spellchange{three available} transition systems on Hierarchical Pointer Networks combined with different fusion functions 
implementations 
and gating mechanisms on ten treebanks from \change{Universal Dependencies}. Only best models on average on dev splits are evaluated on test sets. In Appendix~\ref{sec:standev}, we report the standard deviations over 3 runs on test splits. We use ISO 639-1 codes to represent languages.}
\label{tab:ud}
\end{center}
\end{small}
\end{table*}

\subsection{Results}

Table~\ref{tab:ud} reports LAS on dev splits of \change{Universal Dependencies treebanks} obtained by the proposed bottom-up Hierarchical Pointer Networks with different transition systems and parser configurations (gate and fusion function) and choose 
the best configuration on average to be evaluated on test splits. As shown in the reported results,
the proposed architecture with the three algorithms
improves over the baseline parser in practically all datasets, obtaining higher gains on those languages with a larger amount of long arcs (such as Basque and Chinese). We can also observe that
the simplified fusion functions improve over the adapted and full versions on average, meaning that long-distance dependents are more valuable than last-assigned ones and, in some cases, the usage of the latter harms parsing accuracy. Regarding the novel transition systems, 
the right-to-left parser outperforms other algorithms on languages with a significant predominance of long rightward arcs (such as Arabic),
and
the outside-in algorithm clearly underperforms the other transition systems on average in spite of having access to left and right relevant dependents, which probably means that a sequential human-like strategy is more suitable for parsing natural languages, or that the outside-in order is too complex to learn effectively. Finally, \textsc{Gate1} obtains higher accuracy in general.

\begin{table}[h]
\begin{tabular}{@{\hskip 1pt}lccc@{\hskip 1pt}}
\toprule
\textbf{tran.} & \textbf{$f$} & \textbf{UAS} & \textbf{LAS}  \\
\midrule
\textbf{L2R} & \textbf{l-adapted}  & \textbf{96.03} & \textbf{94.15}  \\
& l-simple  & 96.01 & \textbf{94.15}  \\
\midrule
\textbf{R2L} & r-adapted  & 96.00 & 94.14  \\
& \textbf{r-simple}  & \textbf{96.04} & \textbf{94.16}  \\
\midrule
\textbf{O-I} & full  & 95.96 & 94.10   \\
& \textbf{simple}  &  \textbf{96.00} & \textbf{94.14}   \\
\bottomrule
\end{tabular}

\setlength{\abovecaptionskip}{4pt}
\caption{Accuracy performance on PTB dev splits. We mark in bold the chosen configurations.}
\label{tab:ptbdev}
\end{table}

\begin{table}[h]
\begin{small}
\begin{center}
\centering
\begin{tabular}{@{\hskip 1pt}lcccc@{\hskip 1pt}}
\toprule
 & \multicolumn{2}{@{\hskip 0pt}c@{\hskip 0pt}}{\textbf{PTB}} & \multicolumn{2}{@{\hskip 0pt}c@{\hskip 0pt}}{\textbf{CTB}} \\
\textbf{Parser} & \textbf{UAS} & \textbf{LAS} & \textbf{UAS} & \textbf{LAS} \\
\midrule
\citet{Zhang17} & 94.10 & 91.90 & 87.84 & 86.15 \\
\citet{Ma2017} & 94.88 & 92.96 & 89.05 & 87.74\\
\citet{DozatM17} & 95.74 & 94.08 & 89.30 & 88.23 \\
\citet{li-etal-2018-seq2seq} & 94.11 & 92.08 & 88.78 & 86.23 \\
\citet{ma-etal-2018-stack} & 95.87 & 94.19 & 90.59 & 89.29 \\
\citet{ji-etal-2019-graph}$^\dagger$ & 95.97 & 94.31 & - & - \\
\citet{fernandez-gonzalez-gomez-rodriguez-2019-left} &  96.04 &  94.43 & - & -  \\
\citet{li2019global} & 95.83 & 94.54 & 90.47 & 89.44 \\
\citet{multipointer} & 96.06 & 94.50 & 90.61 & 89.51 \\
\citet{zhang-etal-2020-efficient}$^\dagger$ & 96.14 & 94.49 & - & - \\
\citet{wang2020secondorder} & 95.98 & 94.34 & \textbf{90.81} & 89.57 \\
\textbf{Hier. Ptr. Net. L2R} & \textbf{96.18} & \textbf{94.59} & 90.76 & \textbf{89.67} \\
\textbf{Hier. Ptr. Net. R2L} & 96.14 & 94.53 & 90.72  & 89.62 \\
\textbf{Hier. Ptr. Net. O-I} & 96.07 & 94.48 &  90.64 & 89.50 \\
\hdashline[1pt/1pt]
\textit{+BERT} \\
\ \ \ \citet{li2019global} & 96.44 & 94.63 & 90.89 & 89.73 \\
\ \ \ \citet{li2019global}$^*$ & 96.57 & 95.05 & - & - \\
\ \ \ \citet{mohammadshahi2020recursive}$^*$ & 96.66 & 95.01 & \textbf{92.86} & 91.11 \\
\ \ \ \citet{wang2020secondorder}$^*$ & 96.91 & 95.34 & 92.55 & 91.38 \\
\ \ \ \citet{multipointer} &  96.91 &  95.35 & 92.58 & 91.42  \\
\ \ \ \textbf{Hier. Ptr. Net. L2R} & \textbf{97.05} & 95.47 & 92.70 & 91.50 \\
\ \ \ \textbf{Hier. Ptr. Net. R2L} & 97.01 & \textbf{95.48}  & 92.75 & \textbf{91.62} \\
\ \ \ \textbf{Hier. Ptr. Net. O-I} & 96.95 & 95.36  & 92.65 & 91.47\\
\midrule
\citet{zhou-zhao-2019-head} & 96.09 & \textbf{94.68} & - & - \\
\citet{multipointer} & \textbf{96.25} & 94.64 & \textbf{90.79} & \textbf{89.69} \\
\hdashline[1pt/1pt]
\textit{+BERT} \\
\ \ \ \citet{zhou-zhao-2019-head}$^*$ & \textbf{97.00} & 95.43 & 91.21 & 89.15 \\
\ \ \ \citet{multipointer}  & 96.97 & \textbf{95.46} & \textbf{92.78} & \textbf{91.65} \\
\midrule
\citet{liu-etal-2019-hierarchical} & 96.09 & 95.03 & - & - \\
\bottomrule
\end{tabular}
\centering
\setlength{\abovecaptionskip}{4pt}
\caption{Performance comparison of dependency parsers on PTB and CTB. \spellchange{The} second block gathers approaches that are enhanced with constituent information, \spellchange{and the last block} includes the performance of the top-down transition-based model with Hierarchical Pointer Networks, since only scores with gold PoS tags are reported. $^*$ denotes those models that fine-tune BERT and
those parsers marked with $^\dagger$ report scores based on the best single run on the development set, instead of reporting the average score on the test set over several runs, i.e., instead of averaging to mitigate the effect of random seeds in reported accuracy, they use model selection to choose the most promising seed using the dev set (following this method, our best model \textit{Hier.Ptr.Net L2R} w/o BERT obtains UAS 96.19 LAS 94.61 on PTB).}
\label{tab:penn}
\end{center}
\end{small}
\end{table}

We also compare the proposed approach with the three implemented transition systems
against 
\added{the most recent}
dependency parsers on PTB and CTB. In order to choose the best parser configuration for the Penn Treebanks, we use scores on PTB development splits reported in Table~\ref{tab:ptbdev}. Since \textsc{Gate1} outperforms \textsc{Gate2} in general (as shown for \change{Universal Dependencies treebanks} in Table~\ref{tab:ud}), we simply use \textsc{Gate1} in all experiments and vary the fusion function implementation. Lastly, for BERT augmentations, we directly use simplified fusion functions, significantly reducing training time.
As we can see in Table~\ref{tab:penn}, the 
left-to-right approach
achieves the highest accuracy obtained so far on PTB and CTB test splits with neither 
contextualized word embeddings nor extra constituent information. Regarding BERT augmentations, all transition systems not only outperform the baseline \citep{multipointer}, but also the right-to-left model achieves
the best LAS to date among approaches that use (or fine-tune) BERT (even improving over those enhanced with constituent information on PTB). These results provide some \spellchange{evidence} that our proposed neural architecture has access to some structural information not learnt by the pre-trained language model BERT. \new{Finally, all these improvements in accuracy provided by the proposed neural architecture come without harming the $O(n^2)$ runtime complexity of the original left-to-right transition-based parser, overcoming, therefore, their main competitors that have a higher $O(n^3)$ runtime complexity: the second-order graph-based parsers by \citet{zhang-etal-2020-efficient} and
\citet{wang2020secondorder}.}

\begin{table}[h]
\begin{small}
\begin{center}
\centering
\begin{tabular}{@{\hskip 0pt}lcccccc@{\hskip 0pt}}
\toprule
 & \multicolumn{2}{r}{\textbf{Left2Right}} & \multicolumn{2}{r}{\textbf{Right2Left}} & \multicolumn{2}{r}{\textbf{Outside-in}} \\
 \textbf{Language} & \textbf{all} & \textbf{long} & \textbf{all} & \textbf{long} & \textbf{all} & \textbf{long} \\
\midrule
Arabic  &  10.07 & 1.20 & \textbf{22.20} & \textbf{8.00} & 14.59 & 3.75 \\
Basque  &  6.65 & 2.29 & 5.75 & 2.17	& \textbf{7.10} & \textbf{2.81} \\
Chinese  &   \textbf{16.34} & \textbf{6.17} & 7.99 & 3.83 &	12.60 & 5.34
 \\
English  & \textbf{7.02} & 1.66 & 4.54 & \textbf{2.54} &	6.10 & 2.33
 \\
Finnish  &  \textbf{7.06} & 1.69 & 5.36 & \textbf{2.70} &	6.61 & 2.45
 \\
Hebrew  &  10.80 & 1.69 & \textbf{11.77} & \textbf{5.47} &	11.69 & 3.95
 \\
Italian  &  \textbf{11.15} & 1.74 & 8.97 & \textbf{4.88} & 	10.64 & 3.73
 \\
Korean  &  \textbf{6.94} & \textbf{2.45} & 4.64 & 0.79 &	5.77 & 1.61
 \\
Swedish  & \textbf{10.63} & 2.70 & 7.81 & \textbf{4.18} &	9.61 & 3.71
 \\
Turkish  &  \textbf{5.71} & \textbf{1.85} & 3.46 & 0.74 & 	5.07 & 1.27
\\
\bottomrule
\end{tabular}
\centering
\setlength{\abovecaptionskip}{4pt}
\caption{Number of \spellchange{dependents} and long-range dependents per sentence that are available when processing gold trees in \change{Universal Dependencies} dev splits with each proposed transition system. Note that the number of long-range dependents (with arc lengths $>$ 4) per sentence is notably low since short sentences are also considered for the computation.}
\label{tab:deps}
\end{center}
\end{small}
\end{table}

\change{Finally, while the runtime complexity of the proposed transition-based parsers is quadratic as that of the original sequential approach, the leverage of structural information during decoding comes at a cost in practice. In particular, the left-to-right, right-to-left and outside-in algorithms implemented on the Hierarchical Pointer Network respectively deliver 9.62 sent./s., 9.82 sent./s. and 8.81 sent./s. during decoding time on the test split of the PTB. This means that the original 22.57-sentences-per-second speed of the sequential Pointer Network is approximately halved. Although the current version of our neural architecture is less efficient than the baseline, it is worth mentioning that it was not optimized for speed and, therefore, further enhancements in handling structural information would improve final decoding speed.}

\subsection{Dependent Information Availability}
To further understand the availability of dependent information for each transition system and its impact on parsing performance, 
we report in Table~\ref{tab:deps} the number of dependents per sentence that are accessible for each transition system on \change{Universal Dependencies treebanks}. While, at a first glance, we might think that the outside-in algorithm is the strategy that leverages dependent information the most since it receives both left and right dependents, it seems that, \spellchange{overall} and according to gold trees, the left-to-right parser is the strategy that has access to a larger amount of (in this case, left) \spellchange{dependents, and the right-to-left approach is the one} that uses more long-range (right) dependents (typically present 
in right-branching languages, where rightward dependency arcs tend to be longer). This might explain the good results of the right-to-left approach on \change{Universal Dependencies} datasets on average (since the majority of tested languages have a higher percentage of long rightward arcs and long-range dependents are considered more valuable for reducing error propagation) and the fact that the l-adapted fusion function has a better performance on the left-to-right parser (since this function gathers information about closer dependents and this algorithm has access to a notable amount of dependent information). 
Finally, we can also observe that, as expected, the right-to-left algorithm is more adequate to model languages with a high predominance of rightward arcs such as Hebrew and especially Arabic, where it has access to a notable amount of right dependents not \spellchange{seen} by the other transition systems throughout the parsing process.

\begin{figure*}[h]
\centering
\includegraphics[width=0.45\textwidth]{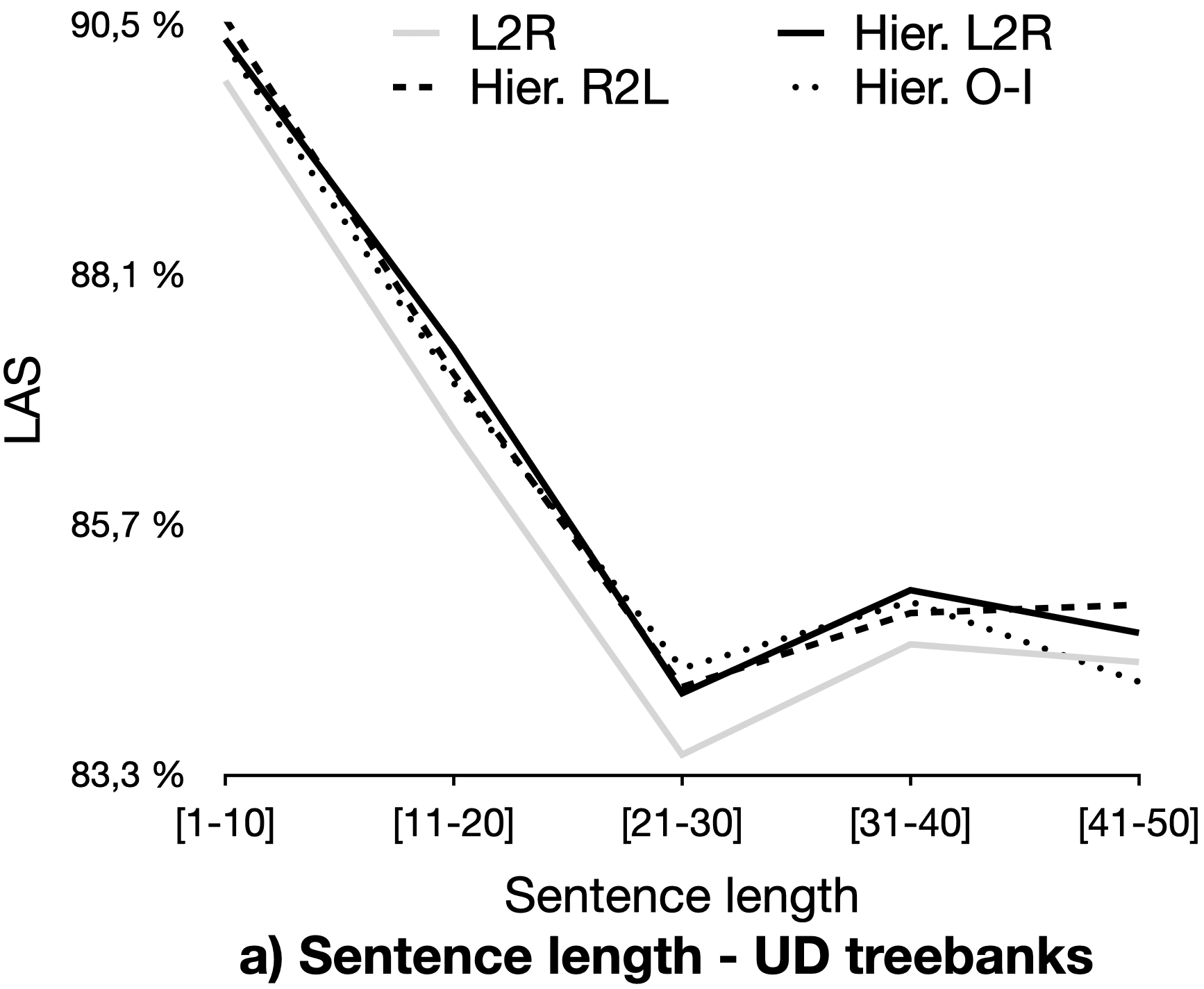}
\vspace{0.5cm}
\includegraphics[width=0.45\textwidth]{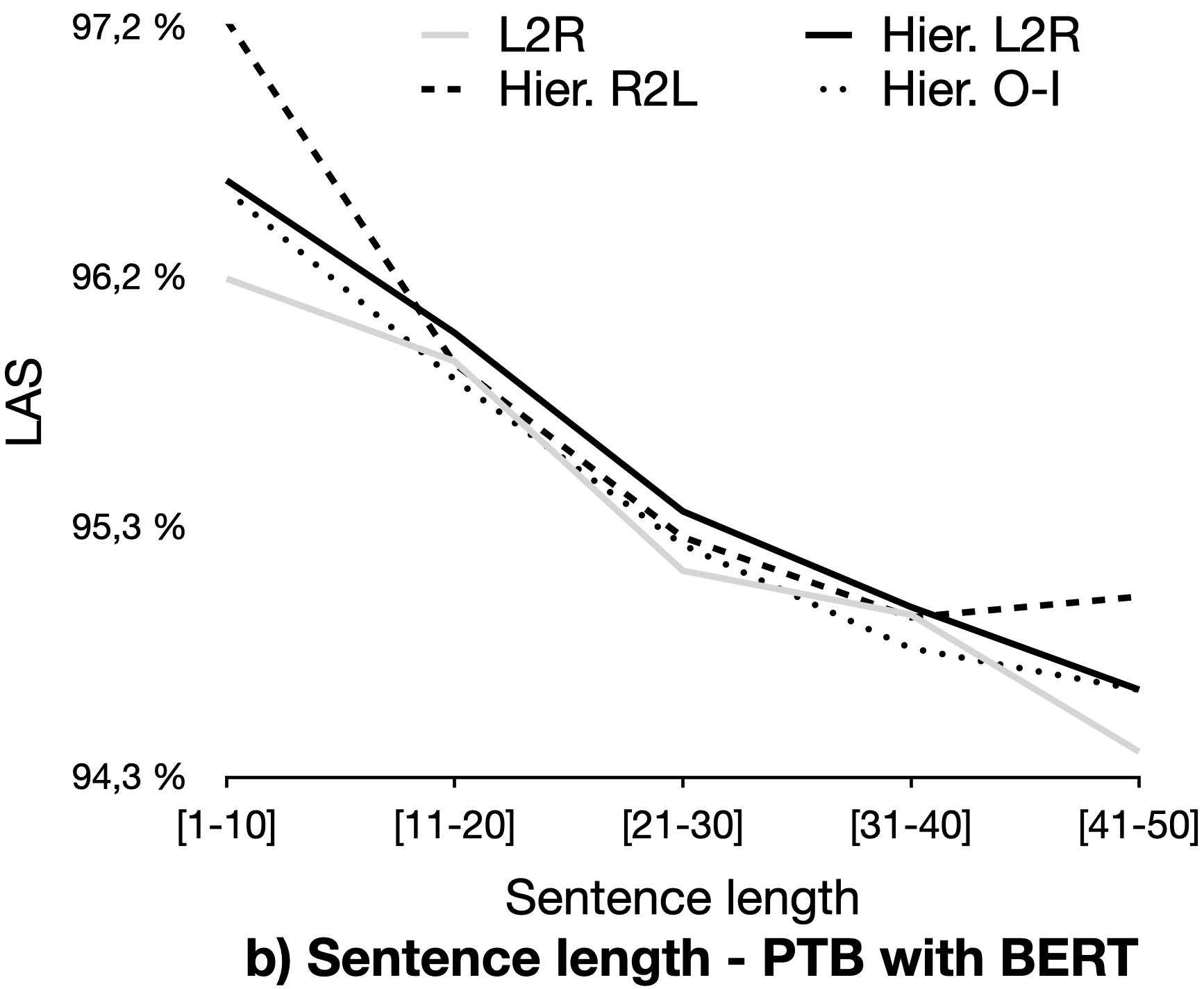}
\includegraphics[width=0.45\textwidth]{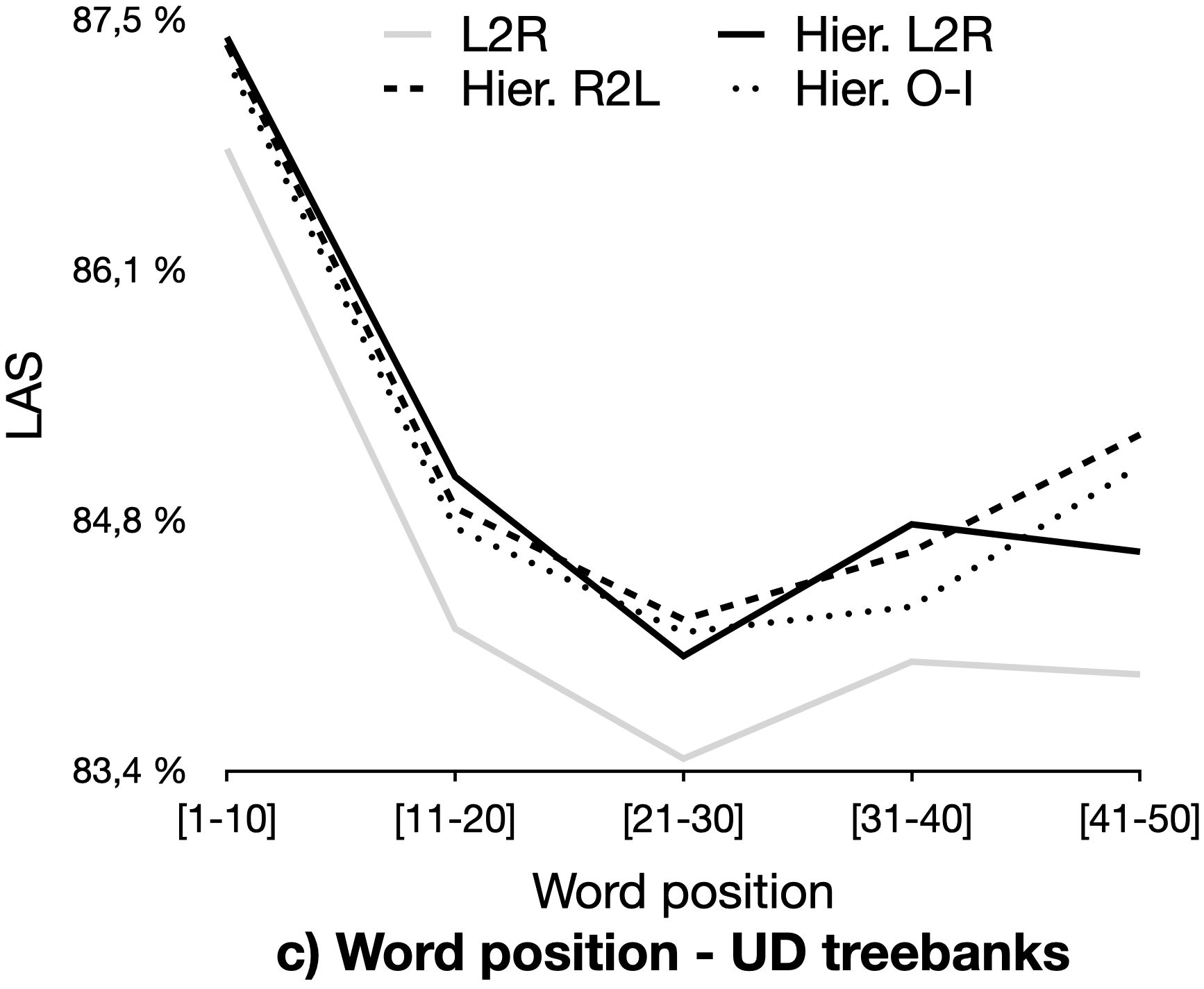}
\vspace{0.5cm}
\includegraphics[width=0.45\textwidth]{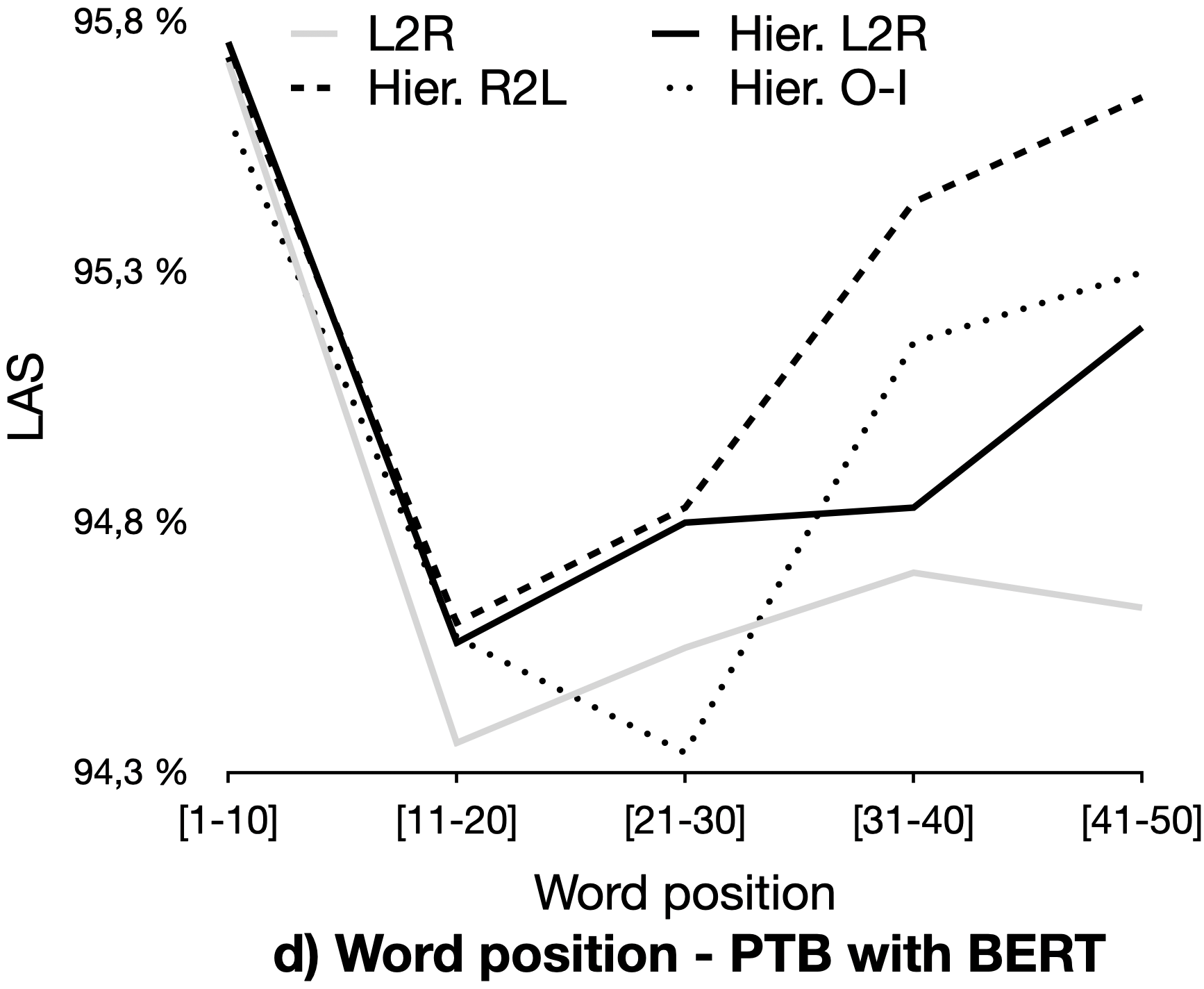}
\caption{Parsing performance of the original left-to-right parser and the proposed variants implemented on Hierarchical Pointer Networks 
relative to sentence length and 
word position.}
\label{fig:analysis}
\end{figure*}

\subsection{Error Analysis}
We study the mitigation of error propagation by
the proposed 
neural 
architecture and characterize errors relative to 
sentence length and word position\footnote{Errors relative to dependency length (as described in \citep{mcdonald-nivre-2007-characterizing}) are no longer an effective method to measure the impact of error propagation in transition systems designed for Pointer Networks (except for the top-down algorithm \citep{ma-etal-2018-stack}), since these require a single \textsc{Shift-Attach}-$p$ transition to directly build any arc (regardless of its length), not involving several decoder states and, therefore,
not directly reflecting the presence of error propagation.}
for a concatenation of all ten \change{Universal Dependencies} datasets\footnote{
In order to prevent data sparsity,
we randomly select \spellchange{a} number of sentences that
approximately gather 10,000 tokens per treebank.} and for the PTB with BERT-based embeddings.

In particular, Figures~\ref{fig:analysis}(a) and (b) show the accuracy relative to sentence lengths of the three 
transition systems 
plus the baseline parser on \change{Universal Dependencies treebanks} 
and 
PTB, respectively. On \change{Universal Dependencies datasets}, we see accuracy gains regardless of sentence length; however, while the improvements tend to be 
higher as sentences are longer, these narrow when the length is larger than 40 (especially affecting the left-to-right and outside-in variants). On PTB, 
the novel approaches not only outperform the baseline on long sentences (as expected), but also surprisingly on the shortest ones. We also note that,
while the right-to-left parser 
obtains the highest performance on the longest sentences (better dealing with error-propagation), the outside-in algorithm is suffering a drop 
on sentences with length greater than 40 on \change{Universal Dependencies treebanks}, 
\spellchange{and on those with length between 31 and 40} on PTB (possibly 
because the information about left and right dependents is exclusively used by this transition system and it tends to be available at final steps 
-- as shown 
in Figure~\ref{fig:algos}(c) --, where it might be difficult to manage when sentences are substantially long and the amount of information \spellchange{is likely to be considerably} large).

The reduction of error propagation can be seen more clearly in
 Figures~\ref{fig:analysis}(c) and (d), where we report the LAS relative to word positions within the sentence. In comparison to the sequential variant, the left-to-right transition system with structured decoding obtains the highest gains in accuracy on attachments made on words at final positions of the sentence (the most affected by error propagation). We can also observe how the right-to-left and outside-in approaches significantly outperform the left-to-right algorithm on words located near the end of the sentence, since these transition systems attach those words in initial steps of the parsing process. Lastly, there is a significant drop in accuracy by the outside-in model in words at middle positions on PTB (from 21 to 30) and, less significant, on \change{Universal Dependencies} datasets (from 31 to 40). \new{This is probably due to the fact that arcs in those positions are created in final steps following the outside-in strategy (thus being the most affected by error-propagation); and, apart from the tight Y axis scale in the plot of Figure~\ref{fig:analysis}(d), the significant drop in PTB can be explained by the fact that the amount of words in absolute positions from 21 to 30 (not necessarily being at the middle of \spellchange{their respective sentences}) is substantially higher than in other languages. }

\section{Conclusions}
\label{sec:conclusion}
\new{We manage to introduce structural knowledge to the sequential decoding of the 
left-to-right dependency parser with Pointer Networks developed by \citet{fernandez-gonzalez-gomez-rodriguez-2019-left}. The resulting neural architecture, \spellchange{named} Bottom-up Hierarchical Pointer Network, is able to leverage relevant information about the focus word’s dependents at each decoding step, instead of just receiving the previous decoder hidden state as input. This information is especially useful when  it comes from long-range dependents attached in the past, notably helping the decoder to take better decisions in current and future steps.} \new{Additionally, we 
implement two novel transition systems that can be added to the proposed neural architecture: an algorithm that parses a sentence in right-to-left order and a transition system that processes it from the outside in.}

\new{We extensively test Bottom-up Hierarchical Pointer Networks on the English and Chinese Penn Treebanks as well as a wide range of languages from different
families, with different \spellchange{degrees of} morphological complexity and with different predominances of long-range dependencies. In our experiments, we prove that they lead to significant accuracy gains 
regardless of the fusion function implementation or gating mechanism; and show that the left-to-right approach achieves the highest accuracy obtained so far on the English and Chinese Penn Treebanks (without contextualized word embeddings) and the right-to-left model achieves the best LAS to date when BERT-based embeddings are available. Apart from the improvements in accuracy, the proposed neural architecture does not penalize the quadratic runtime complexity of the original left-to-right parser. Finally, we additionally undertake a thorough error analysis and 
provide evidence that error propagation (considered the main cause of accuracy loss in transition-based parsing) is mitigated with the presented 
structured
decoding. }


\section*{Acknowledgments}
We acknowledge the European Research Council (ERC), which has funded this research under the European Union’s Horizon 2020 research and innovation programme (FASTPARSE, grant agreement No 714150) and the Horizon Europe research and innovation programme (SALSA, grant agreement No 101100615), ERDF/MICINN-AEI (SCANNER-UDC, PID2020-113230RB-C21), Xunta de Galicia (ED431C 2020/11), and Centro de Investigaci\'on de Galicia ``CITIC'', funded by Xunta de Galicia and the European Union (ERDF - Galicia 2014-2020 Program), by grant ED431G 2019/01.








\printcredits

\bibliographystyle{cas-model2-names}

\bibliography{anthology,main}

\appendix

\section{Standard deviations}
\label{sec:standev}

\begin{table}[h]
\begin{small}
\begin{center}
\centering
\begin{tabular}{@{\hskip 1pt}lcccc@{\hskip 1pt}}
\toprule
 & \multicolumn{2}{@{\hskip 0pt}c@{\hskip 0pt}}{\textbf{PTB}} & \multicolumn{2}{@{\hskip 0pt}c@{\hskip 0pt}}{\textbf{CTB}} \\
\textbf{Parser} & \textbf{UAS} & \textbf{LAS} & \textbf{UAS} & \textbf{LAS} \\
\midrule
\textbf{Hier. Ptr. Net. L2R} & $\pm$0.02 & $\pm$0.03 & $\pm$0.05 & $\pm$0.06 \\
\textbf{Hier. Ptr. Net. R2L} & $\pm$0.04 & $\pm$0.03 & $\pm$0.06 & $\pm$0.08 \\
\textbf{Hier. Ptr. Net. O-I} & $\pm$0.03 & $\pm$0.05 & $\pm$0.04 & $\pm$0.06 \\
\hdashline[1pt/1pt]
\textit{+BERT} \\
\ \ \ \textbf{Hier. Ptr. Net. L2R} & $\pm$0.01 & $\pm$0.02 & $\pm$0.04 & $\pm$0.05 \\
\ \ \ \textbf{Hier. Ptr. Net. R2L} & $\pm$0.03 & $\pm$0.02 & $\pm$0.06 & $\pm$0.05 \\
\ \ \ \textbf{Hier. Ptr. Net. O-I} & $\pm$0.02 & $\pm$0.02 & $\pm$0.03 & $\pm$0.04 \\
\bottomrule
\end{tabular}
\centering
\setlength{\abovecaptionskip}{4pt}
\caption{Standard deviations over 3 runs on test splits for scores reported in Table~\ref{tab:penn}.}
\label{tab:ptbstandev}
\end{center}
\end{small}
\end{table}

\begin{table}[h]
\begin{small}
\begin{center}
\centering
\begin{tabular}{@{\hskip 1pt}lcccccccccccc@{\hskip 1pt}}
\toprule
\textbf{tran.} & \textbf{$f$} & \textbf{gate} & \textbf{ar} & \textbf{en} & \textbf{eu} & \textbf{fi} & \textbf{he} & \textbf{it} & \textbf{ko} & \textbf{sv} & \textbf{tr} & \textbf{zh} \\
\midrule
\textbf{L2R} & - & - & $\pm$0.06 & $\pm$0.01 & $\pm$0.05 & $\pm$0.06 & $\pm$0.14 & $\pm$0.04 & $\pm$0.06 & $\pm$0.03 & $\pm$0.13 & $\pm$0.04 \\
\textbf{L2R} & l-simp. & \textsc{Gate1} & $\pm$0.02 & $\pm$0.04 & $\pm$0.06 & $\pm$0.02 & $\pm$0.08 & $\pm$0.02 & $\pm$0.04 & $\pm$0.04 & $\pm$0.08 & $\pm$0.04 \\
\textbf{R2L} & r-simp. & \textsc{Gate1} & $\pm$0.06 & $\pm$0.04 & $\pm$0.02 & $\pm$0.06 & $\pm$0.12 & $\pm$0.02 & $\pm$0.06 & $\pm$0.05 & $\pm$0.11 & $\pm$0.08
 \\
\textbf{O-I} & simple & \textsc{Gate1} & $\pm$0.02 & $\pm$0.08 & $\pm$0.05 & $\pm$0.04 & $\pm$0.06 & $\pm$0.07 & $\pm$0.09 & $\pm$0.02 & $\pm$0.12 & $\pm$0.08 \\
\bottomrule
\end{tabular}
\centering
\caption{Standard deviations over 3 runs on test splits for scores reported in Table~\ref{tab:ud}.}
\label{tab:standev}
\end{center}
\end{small}
\end{table}

\end{document}